\documentclass[10pt,twocolumn,letterpaper]{article}

\usepackage[ruled,vlined,noresetcount]{algorithm2e}
\usepackage{iccv}
\usepackage{times}
\usepackage{epsfig}
\usepackage{graphicx}
\usepackage{amsmath}
\usepackage{amssymb}

\usepackage{breqn}
\usepackage{adjustbox}
\usepackage{booktabs}
\usepackage{multirow}
\usepackage{dsfont}
\usepackage{cite}
\usepackage{xcolor}

\usepackage[breaklinks=true,bookmarks=false]{hyperref}

\iccvfinalcopy 

\begin{document}

\title{What Synthesis is Missing: Depth Adaptation\\Integrated with Weak Supervision for Indoor Scene Parsing}

\author{Keng-Chi Liu, Yi-Ting Shen, Jan P. Klopp, Liang-Gee Chen\\
National Taiwan University\\
{\tt\small \{calvin89029,dennis45677,kloppjp\}@gmail.com, lgchen@ntu.edu.tw}
}

\maketitle

\begin{abstract}
Scene Parsing is a crucial step to enable autonomous systems to understand and interact with their surroundings. Supervised deep learning methods have made great progress in solving scene parsing problems, however, come at the cost of laborious manual pixel-level annotation.
To alleviate this effort synthetic data as well as weak supervision have both been investigated. Nonetheless, synthetically generated data still suffers from severe domain shift while weak labels are often imprecise. Moreover, most existing works for weakly supervised scene parsing are limited to salient foreground objects. The aim of this work is hence twofold: Exploit synthetic data where feasible and integrate weak supervision where necessary. 
More concretely, we address this goal by utilizing depth as transfer domain because its synthetic-to-real discrepancy is much lower than for color. At the same time, we perform weak localization from easily obtainable image level labels and integrate both using a novel contour-based scheme.
Our approach is implemented as a teacher-student learning framework to solve the transfer learning problem by generating a pseudo ground truth. Using only depth-based adaptation, this approach already outperforms previous transfer learning approaches on the popular indoor scene parsing SUN RGB-D dataset. Our proposed two-stage integration more than halves the gap towards fully supervised methods when compared to previous state-of-the-art in transfer learning.
\end{abstract}

\section{Introduction}

Scene parsing is an important computer vision task aiming at assigning semantic information to the entire image and providing a complete understanding of the scene. 
State-of-the-art scene parsing works \cite{ChenPSA17, ZhaoSQWJ17, LinMSR17, MaSKC17} heavily rely on human labeled pixel-level data, which is expensive and cumbersome to collect.
\begin{figure}[tb]
\begin{center}
\includegraphics[width=\linewidth]{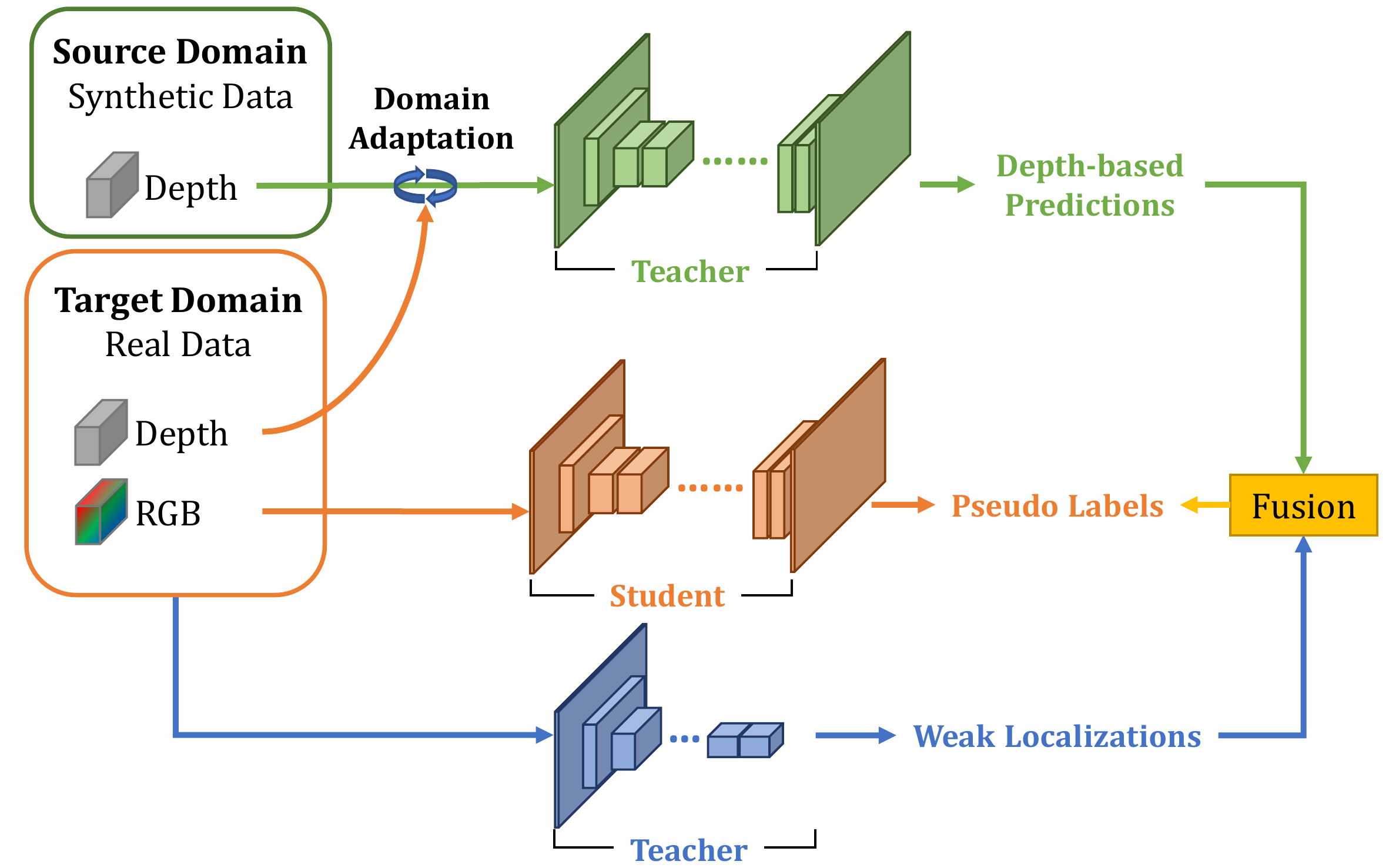}
\end{center}
\vspace{-0.3cm}
\caption{Illustration of our teacher-student framework. The teacher utilize depth as a low domain-shift auxiliary cue. This is fused with weak localization information to generate pseudo labels, which are used to train the student.}
\label{fig:intro}
\end{figure}
To enable computer vision applications without such labeling efforts, two paradigms have been investigated to overcome this issue: unsupervised domain adaptation and weak supervision. Domain adaptation for scene parsing (c.f. \cite{HoffmanTPZISED18}) addresses the problem by transferring from a source domain (simulation) to features that are aligned with target domain (real data) without any labeled target samples. In spite of the progress that has been accomplished in realistic scene rendering and transfer learning approaches, there is still a significant domain discrepancy between real and synthetic imagery, especially in texturing. Weak supervision on the other hand tackles this issue by leveraging weak annotations with lower acquisition costs such as bounding boxes \cite{DaiHS15, PapandreouCMY15, OhBKAFS17, LiAT18}, scribble \cite{LinDJHS16}, points \cite{BearmanRFL16} or even image-level labels \cite{PapandreouCMY15, PathakSLD14, PathakKD15, QiLSZJ16, PinheiroC15, KwakHH17, KolesnikovL16, ShimodaY16, RoyT17, WeiFLCZY17, ChaudhryDT17, abs-1803-10464, abs-1806-04659, abs-1805-04574}. This enables a more cost-effective scaling of training datasets. Nevertheless, for image-level annotations, issues such as lack of boundary information, rare pixels for objects of interest, class co-occurrence and discriminative localization remain tremendous challenges. Moreover, the majority of existing works for weak supervision are only capable of handling salient foreground objects.

In this work, we aim at improving performance by transferring through a path of little domain discrepancy. While RGB images contain rich information, it is difficult to transfer from synthetic to real instances in the RGB domain. Hence, we resort to depth information as an auxiliary cue that can be easily captured and is only used at training time. In the depth domain only the object geometry is of interest, which is easier to accurately synthesize and hence presents less domain shift. Therefore we adapt the depth cue to model sensing artifacts that are encountered in real depth measurements. However, the resulting teacher network is unable to segment all categories properly. Books in a book shelf, for example, do not have a distinctive geometry. To recover such information, we leverage image-level object tags. Such tags are easy to acquire, but do not come with location or boundary information. We hence adapt a weak localization technique to obtain heat maps from RGB images through a network trained solely on these image-level tags. Finally, the localization heat map information is fused with the depth-based predictions to yield a pseudo ground truth, which is in turn used to train the final student network on RGB images only. Fig.~\textcolor{red}{\ref{fig:intro}} illustrates our approach.
Our main contributions can be summarized as follows:

\begin{itemize} 
\item We propose a teacher-student learning procedure to learn scene parsing through low domain shift auxiliary cues and weak domain-specific annotations. The student network is shown to surpass its teacher, leading to 58\% reduction of the gap between state-of-the-art supervised and domain adaptation methods.
\item We are the first to perform depth map adaptation through cycle consistent adversarial networks, utilizing a min-max normalization to ensure proper learning of real depth map noise. It is shown to perform favorably against state-of-the-art domain adaption results on SUN RGB-D \cite{SongLX15}.
\item A two-stage voting mechanism is proposed to integrate cues from depth adaptation and weak localization based on contour maps.
\end{itemize}
In order to facilitate low complexity mobile inference, we furthermore apply complexity reduction techniques to your final model. Related results are presented in the supplementary material as these are not our own contributions. 
\section{Related Work}
\subsection{Domain Adaptation}

Domain adaptation aims at transferring source data to features that are aligned with the target domain so as to generalize the ability of the learned model and improve the performance on the task in target domain without target labels \cite{HoffmanTPZISED18}.
Recently, with the progress made in computer graphics, adaptation between synthetic and real domain has become a popular path for various computer vision tasks.
Several datasets such as SceneNet \cite{McCormacHLD16}, Pbrs \cite{ZhangSYSLJF17} have been proposed for scene parsing.
Unfortunately, severe domain shift is still met by virtue of the difficulties generating photo-realistic imagery.
Therefore, several adaptation methods \cite{HoffmanWYD16, ChenCCTWS17, abs-1711-11556, ZhangDG17, HoffmanTPZISED18} have been proposed to reduce the simulation-to-real gap by means of Generative Adversarial Networks (GAN). 
\cite{HoffmanWYD16} applies techniques of global and category specific adaptation. The global statistics are aligned by using a domain adversarial training technique.
\cite{ChenCCTWS17} extends the approach by not only aligning global statistics but class-specific ones as well. \cite{abs-1711-11556} uses the target guided distillation strategy from \cite{HintonVD15} and spatially-aware adaptation during the training process. \cite{ZhangDG17} applies domain adaptation in a curriculum learning \cite{BengioLCW09} fashion, learning scene parsing from tasks that are less sensitive to the aforementioned domain discrepancy. Moreover, \cite{HoffmanTPZISED18} combines a cycle consistency reconstruction loss as proposed by \cite{ZhuPIE17} with a generative approach to prevent the mapping functions from contradicting each other. 

\subsection{Weak Image-level Supervision}

\begin{figure*}[tb]
\begin{center}
\includegraphics[width=1.0\linewidth]{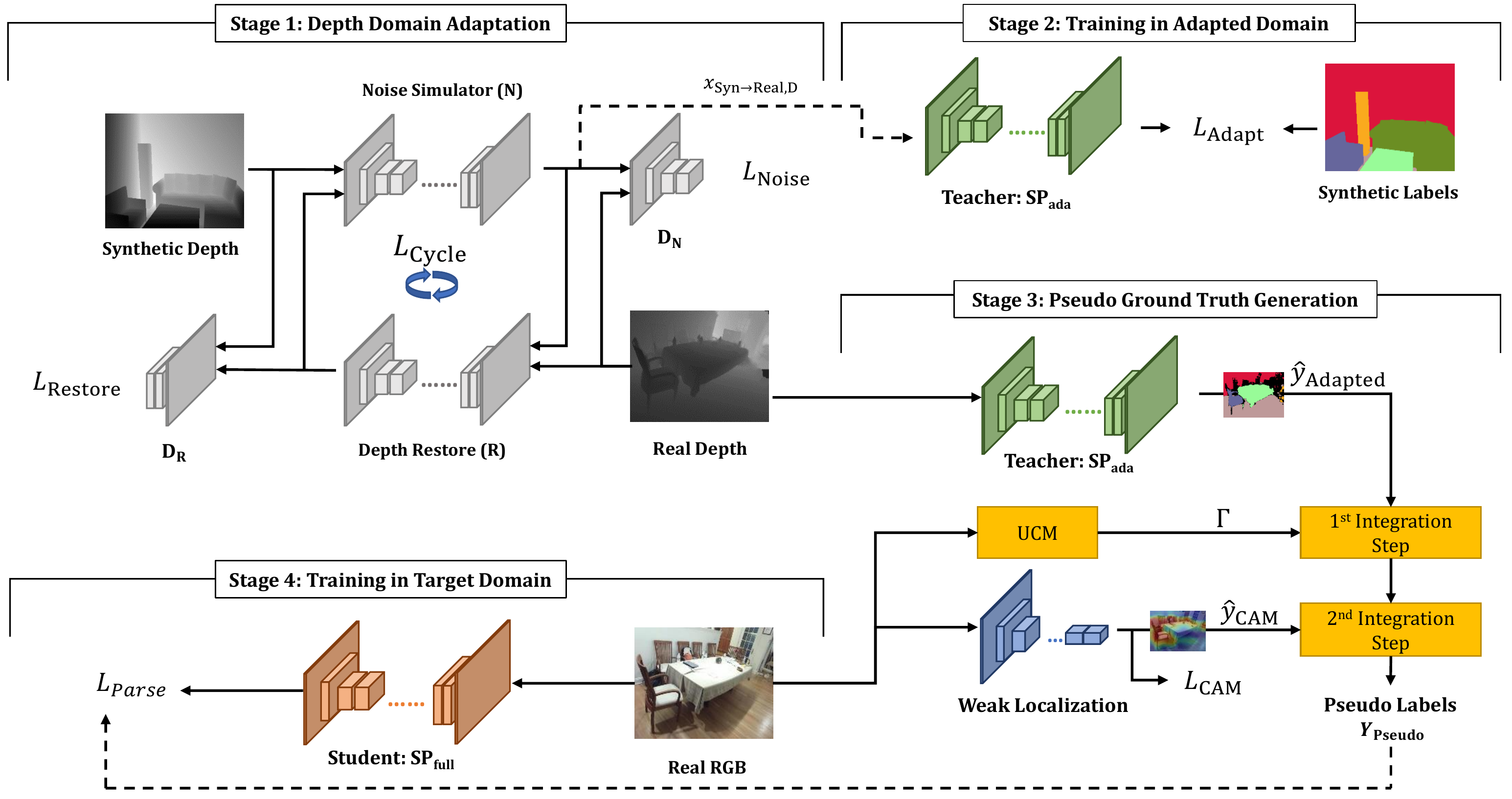}
\end{center}
\vspace{-0.3cm}
\caption{\textbf{Overview of our proposed framework.} A four stage design first adapts synthetic depth maps to appear like real ones. Those adapted depth maps are then used to train a teacher in stage two. Stage three fuses the teacher's predictions with weak localization from class activation maps (CAM) based on contour maps to generate pseudo ground truth. Finally, in stage four, the student network is trained on RGB data using the pseudo labels from the previous stage.}
\label{fig:frame}
\end{figure*}  

Weakly supervised approaches leverage weak annotations that come at lower costs than the original ones.
Since such annotations are efficient to collect, one can build a large-scale dataset for diverse semantic categories with less effort and learn scene parsing in the wild.
Early works mostly applied methods based on graphical models which infer labels for segments with probability relations between images and annotations.
Additionally, class-agnostic cues and post-processing are often used to improve the results.
Among those methods exploiting only weak annotations, learning only from images is the most economical but also challenging one.
Paradigms such as multiple instance learning (MIL) \cite{AndrewsTH02} and self-training \cite{Scudder65a} are often applied.
\cite{PapandreouCMY15} adopt a self-training EM-like procedure, where the model is recursively updated from the results created by itself.
\cite{PathakSLD14} formulates the task as a MIL problem by applying a global max pooling after the CNN to enforce the predictions correspond to positive classes.
Recently, techniques based on discriminative localization \cite{ZhouKLOT16, SelvarajuCDVPB17}, which probe into the contribution of each hidden neuron, are often employed. 
SEC \cite{KolesnikovL16} uses such discriminative localization to indicate a position within the area of a semantic class and expand it to neighboring pixels. However, neural networks tend to focus only on discriminative parts and not on the object as a whole. Hence, works have been focusing on transferring information to the non-discriminative part of objects.
\cite{WeiFLCZY17} obtains improvements by exploiting an adversarial erasing method. Class-agnostic cues are used to obtain shape or instance information in most works that achieve state-of-the-art results \cite{ChaudhryDT17}.
\cite{abs-1806-04659} uses both techniques to mine common object features from the initial localization, expand object regions and consider saliency maps under a Bayesian framework.
\cite{abs-1803-10464} propagates semantic information by a random walk with the affinities predicted by AffinityNet.
\cite{abs-1805-04574} argues that varying dilation rates can effectively promote object localization maps.
Furthermore, most existing works are dedicated to handle multiple salient foreground instances and evaluate on the Pascal VOC dataset\cite{EveringhamEGWWZ15}.
\cite{SalehASPA17} is the only existing work that considers complete scene parsing (background + foreground) with only image-level label by leveraging two-stream deep architecture and heat map loss. However, their result does not perform well compared to other adaptation methods on the Cityscape dataset.

\section{Proposed Method}

In this section, we present the details of our proposed scene parsing framework. Fig.~\textcolor{red}{\ref{fig:frame}} illustrates how it proceeds in four stages: First, we adapt the depth cues from the synthetic into the real domain. Second, we train a teacher network on the adapted synthetic depth cues. Third, by applying the teacher network to the target (real) domain and integrating the generated labels with weak localization over contour maps, we obtain robust pseudo ground truth. Lastly, we train the student network on the target domain RGB input using the constructed pseudo ground truth.

\subsection{Depth Domain Adaptation}
Our objective is to transfer label information from synthetic data $X_{\text{Syn}}=\{X_{\text{Syn,D}},X_{\text{Syn,RGB}}\}$ to the real domain $X_{\text{Real}}=\{X_{\text{Real,D}},X_{\text{Real,RGB}}\}$ while only using depth cues due to their smaller domain gap when compared to RGB. While it is possible to train on synthetic depth data directly, the domain gap still leads to noticeable performance degradation when evaluating on the target (real) domain. Hence, transforming the depth data from the source (synthetic) into the target domain would be beneficial to the later pseudo ground truth generation. This is an unsupervised adaptation problem where only unaligned data from the source and the target domain is available, as only $Y_{\text{Syn}}$ can be accessed while $Y_{\text{Real}}$ cannot. We follow similar adversarial adaptation approaches and learn generators as mappings across these domains (see stage 1 in fig.~\textcolor{red}{\ref{fig:frame}}). In such a setting, discriminators are employed to enforce similarity between the domain mapping and the respective target domain. This alleviates the need for alignment between both domains. In order to construct the sensor noise model (i.e. synthetic to real domain) correctly, we introduce a min-max normalization $\eta$ for depth images:
\begin{dmath}
\eta(I) = 2\times \left(\frac{I-\min(I)}{\max(I)-\min(I)}-\frac{1}{2}\right).
\label{eq:m}
\end{dmath}
By normalizing depth values to lie in the interval $\left [ -1,1 \right ]$ rather than learning in the absolute scale directly, we avoid scale shifting caused by distribution differences among datasets. Additionally, this approach prevents the depth amplitude distribution from becoming the main judging criterion for the discriminator, thereby in turn learning a better sensor noise model. We introduce the sensor noise model $N$ which maps data from the synthetic to the real domain for the purpose of adding realistic noise to synthetic clean samples. It will be optimized to prevent the discriminator $D_{N}$ from distinguishing between mapped and real depth data. The discriminator, on the other hand, tries to differentiate real noisy data from the mapped ones.
We express this objective as:
\begin{dmath}
L_{\text{Noise}}(N,D_{N},X_{\text{Syn,D}},X_{\text{Real,D}})=\mathbb{E}_{x_{t}\sim\ X_{\text{Real,D}}}[log\ D_{N}(\eta(x_{t}))]+\mathbb{E}_{x_{s}\sim\ X_{\text{Syn,D}}}[log\ (1-D_{N}(\eta(N(\eta(x_{s})))))],
\label{eq:gan_n}
\end{dmath}
where Eq.~\textcolor{red}{\ref{eq:gan_n}} ensures that $N$ produces convincing sensor-like noisy samples given synthetic clean samples $X_{\text{Syn},D}$.
Nonetheless, existing works indicate that networks optimizing such objectives are often unstable, mainly because $L_{\text{Noise}}$ does not consider preservation of the original content. Hence a cycle-consistency constraint \cite{ZhuPIE17} is imposed on our adaptation procedure. For that purpose, the restoration model $R$ is introduced to map the sensor-like depth map back to the synthetic clean domain, optimising a similar min-max adversarial loss:
\begin{dmath}
L_{\text{Restore}}(R,D_{R},X_{\text{Real,D}},X_{\text{Syn,D}})=\mathbb{E}_{x_{t}\sim\ X_{\text{Syn,D}}}[log\ D_{R}(\eta(x_{t}))]+\mathbb{E}_{x_{s}\sim\ X_{\text{Real,D}}}[log\ (1-D_{R}(\eta(N(\eta(x_{s})))))].
\end{dmath}
In contrast to the noise simulator $N$, the restorer $R$ performs tasks such as hole filling and denoising. More details on how this is accomplished along with qualitative results can be found in the supplementary material. Moreover, an L1 penalty is imposed on samples mapped twice so as to reach to their original domain again, e.g. mapping a synthetic sample to the sensor-like depth domain and back to the synthetic domain. This is referred to as the min-max cycle-consistency loss:
\begin{dmath}
L_{\text{Cycle}}(N,R)=\mathbb{E}_{x_{s}\sim\ X_{\text{Syn,D}}}[\left \| R(\eta(N(\eta(x_{s})))) -\eta(x_{s})\right \|]+\mathbb{E}_{x_{t}\sim\ X_{\text{Real,D}}}[\left \| N(\eta(R(\eta(x_{t})))) -\eta(x_{t})\right \|].
\end{dmath}
These three loss functions form our complete objective:
\begin{dmath}
L(N,R,D_{N},D_{R}) = L_{\text{Noise}}(N,D_{N},X_{\text{Syn,D}},X_{\text{Real,D}})+L_{\text{Restore}}(R,D_{R},X_{\text{Real,D}},X_{\text{Syn,D}})+L_{\text{Cycle}}(N,R).
\end{dmath}
Finally, we train the two autoencoders $N, R$ and their respective discriminators, $D_N$ and $D_R$, jointly by solving the following optimization problem:
\begin{dmath}
N, R = \arg \min_{N,R} \max_{D_{N},D_{R}}L(N,R,D_{N},D_{R}).
\end{dmath}

\subsection{Training in Adapted Domain}
The ability to simulate noise on synthetic training samples enables us in stage two to train a scene parsing model $\mathit{SP}_{\text{ada}}$ using the noisy synthetic training samples that mimic the real training samples, denoted $X_{\text{Syn}\rightarrow \text{Real,D}}=\left\{ N(\eta(x_{\text{Syn,D}}))\forall x_{\text{Syn,D}}\in X_{\text{Syn,D}} \right\}$, and the corresponding labels $Y_{\text{Syn}}$. We train the model by minimizing a pixel-wise multinomial logistic regression loss. Additionally, to prevent overfitting towards an unbalanced class distribution, we apply the class balancing strategy proposed in \cite{PaszkeCKC16}. Formally, the weighted negative log likelihood loss between the prediction and synthetic ground truths for pixel $i$ from a sample $x_{\text{Syn}\rightarrow \text{Real,D}}$ can be written as
\begin{dmath}
L_{\text{Adapt},i} = - \sum_{c\in C }w_{c} y_{i,c}\log\left(\frac{e^{p_{i,c}}}{\sum _{c'\in C}e^{p_{i,c'}}}\right),
\end{dmath}
where $p_{i,c}$ is the prediction made by $\mathit{SP}_{\text{ada}}$, $y_{i,c}$ the ground truth label, $C$ the set of classes with weights $w_{c}$.

\subsection{Pseudo Ground Truth Generation}
The third stage utilizes the predictions made by teacher model $\mathit{SP}_{\text{ada}}$ on real depth data and proceeds to generate pseudo ground truth labels $Y_{\text{Pseudo}}$.
\subsubsection{Weak Localization}
Experiments on depth-only input reveal that the performance of $\mathit{SP}_{\text{ada}}$ is still insufficient for certain categories, e.g. books, as their geometry is not distinctive enough. In attempts to remedy this by adapting a model in the RGB domain, we observed a performance drop nonetheless. This may be due to the domain shift between synthetic and real textures, as a result from the difficulties to model and render certain textures accurately in an automated fashion. Consequently, we propose to utilize weak supervision base on real RGB data as a separate cue for fine-tuning to the object appearance in the target domain. To avoid high labeling costs, we only use image-level tags extracted from SUN RGB-D, without location or boundary information. We generate localization cues by leveraging a CNN that is trained for image classification with a global average pooling layer (GAP) as proposed by \cite{ZhouKLOT16}. However, the resulting class activity maps (CAM) $\hat{Y}_{\text{CAM}}$ are imprecise, \cite{KolesnikovL16} even noted that networks trained with a final GAP overestimate the response region. Hence, we add a $2\times 2$ max pooling layer before the GAP to extract key information and prevent the GAP from overestimation.
\begin{figure}[tb!]
	\begin{center}
		\includegraphics[width=0.8\linewidth]{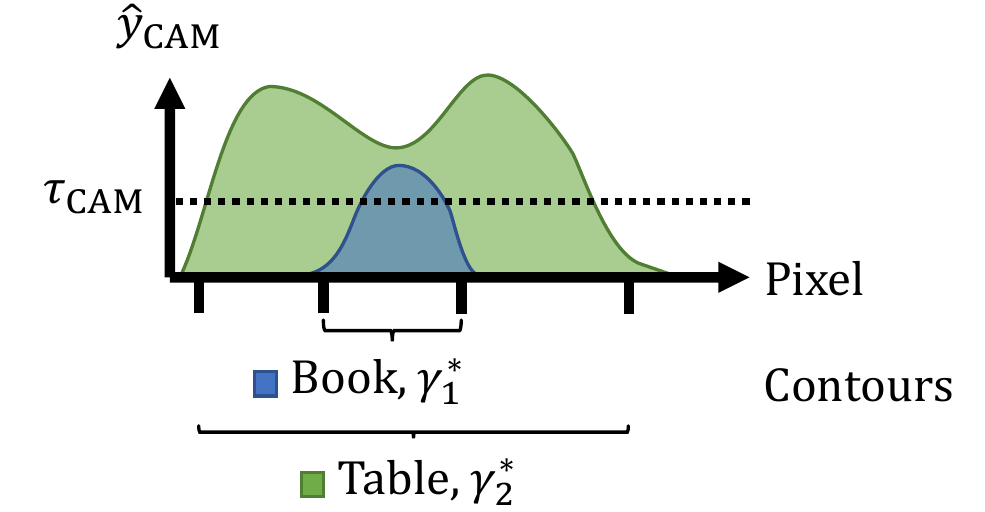}
	\end{center}
	\vspace{-0.3cm}
	\caption{Typical response of the localization heat map for small objects placed on or in larger ones. The response area's size (exceeding $\tau_{\text{CAM}}$) is hence used as a decision criterion if there are several confident predictions for a single contour.}
	\label{fig:integration_step2}
\end{figure}  
\subsubsection{Cue Integration}
To integrate the depth-based predictions  $\hat{Y}_{\text{Adapted}}=SP_{\text{ada}}(X_{\text{\text{Real}},\text{D}})$ and the weak localizations $\hat{Y}_{\text{CAM}}$, we propose a two-stage integration mechanism.
Our objective is to generate pseudo labels $\hat{Y}_{\text{Pseudo}}$ where we trade coverage for confidence: We prefer learning from fewer but more confident pseudo labels. This trade-off is category-related, different categories have different coverage-confidence profiles that need to be accommodated.
We utilize Ultrametric Contour Maps (UCM) \cite{GuptaGAM14}, a hierarchical representation of the image boundaries, to infer pseudo labels over segments $\gamma_k\in\Gamma$ of the image. We only take those contours into account that exceed a confidence threshold $\tau_{\text{UCM}}$, denoting them $\gamma_k^\ast\in\Gamma^\ast$. 


\paragraph{First Integration Step}
The first step adds information about the observed geometry to the contours $\gamma_k^\ast\in\Gamma^\ast$ by analyzing the depth-based predictions $\hat{Y}_{\text{Adapted}}$ within each contour. In order to remove low confidence labels from $\hat{Y}_{\text{Adapted}}$, we first apply a Softmax and threshold the result against $\tau_{\text{Adapted}}=0.6$, resulting in $\hat{Y}_{\text{Adapted}}^\ast$. $\tau_{\text{Adapted}}$ was chosen so as to balance accuracy and coverage. We then turn to the histogram $H(\gamma_k^\ast, \hat{y}_{\text{Adapted}}^\ast)=\left\{h_{c,k}\right\}_{c\in\text{Categories}}$ of predicted categories within each contour $\gamma_k^\ast$. Taking a simple maximum likelihood approach, we select the category with the largest histogram value to be the prediction of the first integration step, i.e. $\hat{y}_{\text{Step 1},k}=\arg\max_c h_{c,k}$ for each contour.

\begin{algorithm}[htb]
	\SetAlgoLined
	\KwResult{Pseudo Labels $y_{\text{Pseudo}}$}
	$O_{\text{Scene Bounds}}=\{\text{Ceil},\text{Floor},\text{Wall}\}$\;
	$O_{\text{Small}}=\{\text{Books},\text{Paint}\}$\;
	\ForEach{Category $c\in C$}{
		$A_c=\sum_{\text{Pixel i}}\hat{y}_{\text{CAM},i,c}>\tau_{\text{CAM}}$\;
	}
	\ForEach{Contour $\gamma_k^\ast\in\Gamma^\ast$}{
		// Add Maximum Likelihood candidate
		$P_k=\left\{\arg\max_{c\in C} \sum_{i\in \gamma_k^\ast} \hat{y}_{\text{CAM},c,i}\right\}\cup O_{\text{Small}}$\;
		// Compute confidence features\\
		$E_k=\emptyset$\;
		\ForEach{Category $c\in P_k$}{
			$p_{k,c}=\max_{i\in\gamma_k^\ast} \hat{y}_{\text{CAM},i,c}$\;
			$r_{k,c}=\frac{\sum_{i\in \gamma_k^\ast}\hat{y}_{\text{CAM},i,c}>\tau_{\text{CAM}}}{\#\gamma_k^\ast}$\;
			// Check if thresholds are met \\			
			\Switch{$y_{\text{Step 1},k}$}{
				\Case{is "Unknown"}{
					$\tau_{p}=\tau_{p,\text{Unknown}}$\;
					$\tau_{r}=\tau_{r,\text{Unknown}}$\;
				}
				\Case{is in $O_{\text{Scene Bounds}}$}{
					$\tau_{p}=\tau_{p,\text{Scene Bounds}}$\;
					$\tau_{r}=\tau_{r,\text{Scene Bounds}}$\;
				}
				\Other{
					$\tau_{p}=\tau_{p,\text{Other}}$\;
					$\tau_{r}=\tau_{r,\text{Other}}$\;
				}
			}
			\If{$p_{k,c}>\tau_p$ and $r_{k,c}>\tau_r$}
			{
				$E_k=E_k\cup\{c\}$\;
			}
		}
		\eIf{$E_k=\emptyset$}{
			$\forall i\in \gamma_k^\ast: y_{\text{Pseudo},i} =y_{\text{Step 1},k}$
		}{
			$\forall i\in \gamma_k^\ast: y_{\text{Pseudo},i} =\mathop{\arg\min}_{c \in E_k} A_c$\;
		}
	}
	\label{alg:integration_step2}
	\caption{Second Integration Step.}
\end{algorithm}
\paragraph{Second Integration Step}  
The second integration step decides whether the localization heat maps $\hat{y}_{\text{CAM}}$ provide a more confident prediction than the contours' $\hat{y}_{\text{Step 1},k}$. Algorithm~\textcolor{red}{\ref{alg:integration_step2}} gives the formal specification. From $\hat{y}_{\text{CAM}}$ we first generate a proposal set of possible classes $P_k$ for each contour, comprising the most activated class in the heat map and a set of small objects. Next, we use peak activation $p_{k,c}$ and response rate $r_{k,c}$ to determine which of the proposals is confident enough to replace the estimate from step one, forming the electable set $E_k$. If this set is empty, i.e. there are no confident localizations, we resort to the result from step one. Otherwise we may have to decide which of several confident localizations is correct, where we take the one with the smallest response area $A_c$. This way, we avoid neglecting small objects that overlap with larger ones (e.g. books on a table as shown in Fig.~\textcolor{red}{\ref{fig:integration_step2}}). All thresholds $\tau $ are tuned on 30 random samples of our training set to avoid “human learning” on the dataset.


\begin{table*}[!htbp]
\centering
\caption{Ablation study of minmax normalization for depth adaptation. Results reported are from the SUN RGB-D validation set. Best values are highlighted in bold font.}
\label{tab:normcomp}
\resizebox{\textwidth}{!}{
\begin{tabular}{@{}cccccccccccccc@{}}
\toprule
 & bed   & books & ceil  & chair & floor & furn. & objs. & paint & sofa  & table & tv   & wall  & \begin{tabular}[c]{@{}c@{}}mIoU\\(w/o windows)\end{tabular} \\ \midrule
\begin{tabular}[c]{@{}c@{}}Ours Depth \\(Raw, w/o minmax normalization)\end{tabular} & 27.85 & 0.00  & 28.36 & 26.37 & 72.29 & 24.84 & 10.91 & 4.13  & 23.28 & 34.21 & \textbf{6.23} & 58.78 & 26.44  \\
Ours Depth (Raw)                                                                                   & \textbf{40.20} & 0.00  & \textbf{33.77} & \textbf{31.21} & \textbf{72.30} & \textbf{30.06} & \textbf{11.61} & \textbf{13.02} & \textbf{31.75} & \textbf{40.13} & 4.49 & \textbf{62.81} & \textbf{30.95}                                                                      \\ \bottomrule
\end{tabular}
}
\end{table*}

\begin{table*}[tb]
\centering
\caption{Ablation study of sensor noise simulation. These results were reported on both inpainted and raw SUN RGB-D validation set.}
\label{tab:mixcomp}
\resizebox{\textwidth}{!}{
\begin{tabular}{@{}cccccccccccccc@{}}
\toprule
 & bed       & books     & ceil    & chair    & floor  & furn.  & objs. & paint & sofa & table & tv & wall & \begin{tabular}[c]{@{}c@{}}mIoU\\ (w/o windows)\end{tabular} \\ \midrule
\multicolumn{14}{c}{\textit{\textbf{Inpainted}}}     \\ \midrule
Syn Depth               & \multicolumn{1}{l}{33.06} & \multicolumn{1}{l}{0.00} & \multicolumn{1}{l}{25.86} & \multicolumn{1}{l}{24.42} & \multicolumn{1}{l}{76.22} & \multicolumn{1}{l}{26.70} & \multicolumn{1}{l}{9.85}  & \multicolumn{1}{l}{9.74}  & \multicolumn{1}{l}{26.22} & \multicolumn{1}{l}{38.70} & \multicolumn{1}{l}{6.36} & \multicolumn{1}{l}{\textbf{63.91}} & 28.42\\
\cite{BohgRHS14}+Syn Depth & \multicolumn{1}{l}{38.55} & \multicolumn{1}{l}{0.00} & \multicolumn{1}{l}{\textbf{37.60}} & \multicolumn{1}{l}{41.21} & \multicolumn{1}{l}{78.25} & \multicolumn{1}{l}{28.28} & \multicolumn{1}{l}{12.80} & \multicolumn{1}{l}{\textbf{16.26}} & \multicolumn{1}{l}{29.41} & \multicolumn{1}{l}{39.71} & \multicolumn{1}{l}{5.85} & \multicolumn{1}{l}{63.34} & 32.61   \\
Ours Depth   & \multicolumn{1}{l}{\textbf{49.04}} & \multicolumn{1}{l}{0.00} & \multicolumn{1}{l}{35.75} & \multicolumn{1}{l}{\textbf{41.40}} & \multicolumn{1}{l}{\textbf{79.55}} & \multicolumn{1}{l}{\textbf{31.44}} & \multicolumn{1}{l}{\textbf{14.68}} & \multicolumn{1}{l}{14.63} & \multicolumn{1}{l}{\textbf{38.51}} & \multicolumn{1}{l}{\textbf{43.73}} & \multicolumn{1}{l}{\textbf{7.78}} & \multicolumn{1}{l}{61.83} & \textbf{34.86}  \\ \midrule
\multicolumn{14}{c}{\textit{\textbf{Raw}}} \\ \midrule
Syn Depth               & 25.92                     & 0.00                     & 31.37                     & 18.97                     & 54.30                     & 22.25                     & 6.95                      & 8.22                      & 19.40                     & 29.24                     & 2.96                     & 47.02                     & 22.22                                                    \\
\cite{BohgRHS14} Depth               & 30.31                     & 0.00                     & 33.54                     & 22.89                     & \textbf{72.40}            & 26.43                     & 11.11                     & 13.01                     & 25.54                     & 36.34                     & \textbf{4.57}      & 61.12                     & 28.11                                                       \\
Ours Depth(Raw)        & \textbf{40.20}               & 0.00                     & \textbf{33.77}         & \textbf{31.21}    & 72.30                     & \textbf{30.06} & \textbf{11.61}  & \textbf{13.02} & \textbf{31.75} & \textbf{40.13}  & 4.49                     & \textbf{62.81}   & \textbf{30.95}                                                        \\ \bottomrule
\end{tabular}
}
\end{table*}

\begin{table*}[tb]
\centering
\caption{Influence of cues and voting mechanism in each stage. These results were obtained on SUN RGB-D validation set.}
\label{tab:ablation}\vspace{0.0cm} 
\resizebox{\textwidth}{!}{
\begin{tabular}{@{}ccccccccccccccccc@{}}
\toprule
           & Input                   & Training Label   & bed   & books & ceil  & chair & floor & furn. & objs. & paint & sofa  & table & tv    & wall  & window & \textbf{mIOU}  \\ \midrule

Ours Depth & Depth              & $Y_{\text{Syn}}$      & 49.04 & 0.00  & 35.75 & 41.40 & 79.55 & 31.44 & 14.68 & 14.63 & 38.51 & 43.73 & 7.78  & 61.83 & 0.91   & 32.25 \\
Ours(1st stage only) & RGB                & $\hat{Y}_{\text{Step 1}}$ & \textbf{54.93} & 0.00  & \textbf{53.12} & 47.50 & 79.64 & 35.77 & 15.99 & 0.00  & 40.39 & \textbf{48.89} & 16.07 & 64.82 & 0.65   & 35.21 \\
Ours(2nd stage only) & RGB                & UCM+$\hat{Y}_{\text{CAM}}$  & 27.71 & 12.87 & 16.13 & 36.19 & 29.17 & 13.12 & 12.95 & 20.15 & 34.56 & 31.27 & 7.81  & 50.72 & \textbf{44.99}  & 25.97 \\
Ours(Full) & RGB                & $Y_{\text{Pseudo}}$     & 52.06 & \textbf{23.52} & 50.03 & \textbf{49.44} & \textbf{81.00} & \textbf{36.39} & \textbf{25.17} & \textbf{28.09} & \textbf{44.64} & 47.88 & \textbf{19.68} & \textbf{69.69} & 38.25  & \textbf{43.53} \\ \bottomrule
\end{tabular}
}
\end{table*}

\begin{table*}[tb]
\centering
\caption{Comparison of pseudo labels $Y_{\text{Pseudo}}$ to our final model. Quantities labeled "effective" refer to the original quantity multiplied by the cover ratio, thereby taking only valid pixels into account for a more accurate comparison. Those labeled @$Y_{\text{Pseudo}}$ are evaluated only on those pixels where pseudo labels $Y_{\text{Pseudo}}$ are available. Evaluations are conducted on the SUN RGB-D dataset. GA refers to the Global Accuracy over all pixels.}
\label{tab:compst}
\footnotesize
\begin{tabular}{@{}lcccccccc@{}}
	\toprule
	Predictions                               & Dataset partition & Cover ratio & GA    & GA@$Y_{\text{Pseudo}}$ & Effective GA & mIoU  & mIoU@$Y_{\text{Pseudo}}$ & Effective mIoU \\ \midrule
	$Y_{\text{Pseudo}}$                       & Training          & 72.77       & 80.86 & 80.86                  & 58.84        & 56.97 & 56.97                    & 41.64          \\
	$\mathit{SP}_{\text{full}}$                        & Training          & 100         & 75.89 & 80.91                  & 75.89        & 49.46 & 56.74                    & 49.46          \\
	$\mathit{SP}_{\text{full}}$ (incl. UCM refinement) & Training          & 97.73       & 76.81 & 81.29                  & 75.07        & 50.81 & 57.52                    & 49.66          \\
	$\mathit{SP}_{\text{full}}$                        & Validation        & 100         & 73.64 & -                      & 73.64        & 43.53 & -                        & 43.53          \\ \bottomrule
\end{tabular}
\end{table*}

\begin{table*}[tb]
\centering
\caption{Comparison of our approach to state-of-the-art domain adaptation and fully-supervised methods. Results are obtained on the SUN RGB-D validation set.}
\label{tab:compalg}\vspace{0.3cm} 
\resizebox{\textwidth}{!}{
\begin{tabular}{@{}cccccccccccccccccrr@{}}
\toprule
\multirow{2}{*}{Method}                                                & \multicolumn{3}{c}{Dataset}                                       & \multirow{2}{*}{bed} & \multirow{2}{*}{books} & \multirow{2}{*}{ceil} & \multirow{2}{*}{chair} & \multirow{2}{*}{floor} & \multirow{2}{*}{furn.} & \multirow{2}{*}{objs.} & \multirow{2}{*}{paint} & \multirow{2}{*}{sofa} & \multirow{2}{*}{table} & \multirow{2}{*}{tv} & \multirow{2}{*}{wall} & \multirow{2}{*}{window} & \multirow{2}{*}{mIOU} & \multirow{2}{*}{\begin{tabular}[c]{@{}c@{}}mIOU\\drop (rel.)\end{tabular}}\\
& SUN  & Scene & Pbrs &  &  & &     &   &     &    &  &   &  &    &  &   &    &   \\ \midrule
Supervised\cite{RomeraABA18}      & \begin{tabular}[c]{@{}c@{}}${\surd}$\\ (full)\end{tabular} &       &      & 62.46                & 26.07                  & 67.54                 & 62.52                  & 85.68                  & 47.10                  & 38.43                  & 43.15                  & 49.72                 & 59.33                  & 40.49               & 76.92                 & 54.12                   & 54.89            & -     \\
NADA\cite{ZhangSYSLJF17}                                                               &                                                    &       & ${\surd}$    & 22.13                & 0.00                   & 23.42                 & 40.08                  & 69.58                  & 23.70                  & 10.34                  & 5.05                   & 36.38                 & 21.90                  & 8.97                & 57.15                 & 23.27                   & 26.31      & -52.07\%        \\
CYCADA\cite{HoffmanTPZISED18}                                                            &                                                    &       & ${\surd}$    & 28.22                & 0.00                   & 24.39                 & 39.57                  & 68.45                  & 23.51                  & 12.61                  & 15.42                  & 39.00                 & 16.65                  & 13.74               & 59.12                 & 34.95                   & 28.90       & -47.35\%           \\
Ours Depth                                                          &                                                    &       & ${\surd}$    & 48.11                & 0.00                   & 22.24                 & 39.99                  & 77.18                  & 27.59                  & 13.92                  & 12.01                  & 39.35                 & 39.32                  & 6.34                & 59.08                 & 0.00                    & 29.24         & -46.73\%        \\
Ours Depth                                                          &                                                    & ${\surd}$     & ${\surd}$    & 49.04                & 0.00                   & 35.75                 & 41.40                  & 79.55                  & 31.44                  & 14.68                  & 14.63                  & 38.51                 & 43.73                  & 7.78                & 61.83                 & 0.91                    & 32.25      & -41.25\%         \\
Ours (Full)                                                           & \begin{tabular}[c]{@{}c@{}}${\surd}$\\ (weak)\end{tabular} & ${\surd}$     & ${\surd}$    & 52.06                & \textbf{23.52}                  & \textbf{50.03}         & 49.44                  & 81.00                  & 36.39                  & \textbf{25.17}       & 28.09                  & 44.64                 & 47.88                  & \textbf{19.68}    & 69.69                 & 38.25                   & 43.53      & -20.70\%          \\
\begin{tabular}[c]{@{}c@{}}Ours (Full\\ +UCM Refinement)\end{tabular} & \begin{tabular}[c]{@{}c@{}}${\surd}$\\ (weak)\end{tabular} & ${\surd}$     & ${\surd}$    & \textbf{54.07}                & 21.94                  & 47.54                 & \textbf{50.37}          & \textbf{81.10}          & \textbf{36.56}       & 24.75                  & \textbf{30.67}         & \textbf{46.23}           & \textbf{49.15}     & 17.76               & \textbf{70.19}      & \textbf{39.00}    & \textbf{43.80}      & \textbf{-20.20\%}             \\ \bottomrule
\end{tabular}
}
\end{table*}

\begin{figure*}[tb]
\begin{adjustbox}{addcode={\begin{minipage}{\textwidth}}{
				\caption{%
      Visualization and comparison of our method.}\label{fig:finalcomp}\end{minipage}}}
\centering
\setlength{\tabcolsep}{1pt}
  \begin{tabular}{@{}cccccccc@{}}
     RGB &  GT &  Supervised &  CYCADA &  Ours-Depth &  Ours-Full &  Ours-Full (+UCM)\\
    \includegraphics[width=0.14\textwidth]{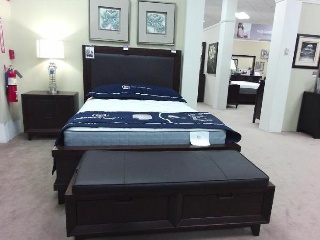} &
    \includegraphics[width=0.14\textwidth]{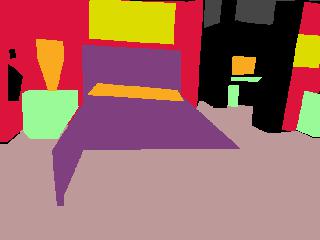} &
    \includegraphics[width=0.14\textwidth]{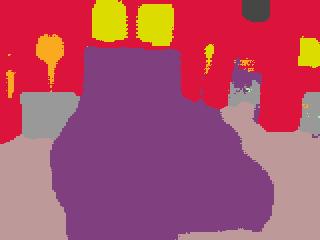} &
    \includegraphics[width=0.14\textwidth]{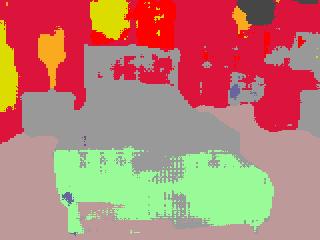} &
    \includegraphics[width=0.14\textwidth]{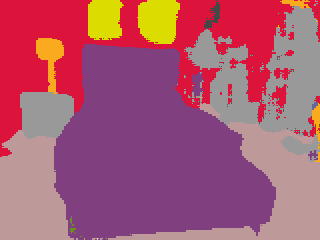} &
    \includegraphics[width=0.14\textwidth]{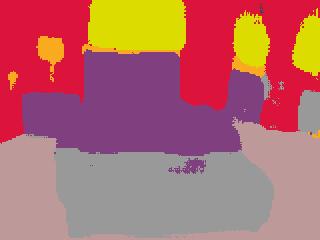} &
    \includegraphics[width=0.14\textwidth]{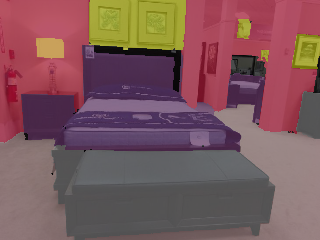} \\
    
    \includegraphics[width=0.14\textwidth]{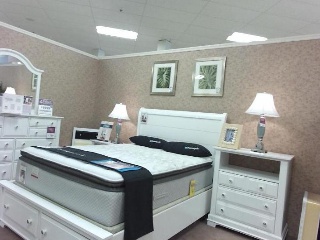} &
    \includegraphics[width=0.14\textwidth]{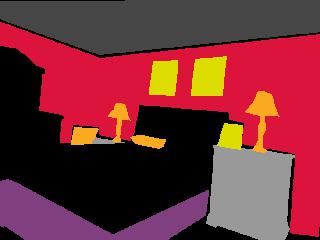} &
    \includegraphics[width=0.14\textwidth]{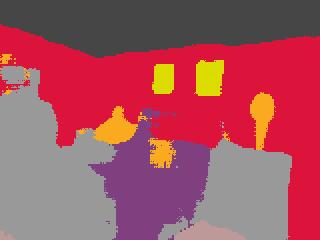} &
    \includegraphics[width=0.14\textwidth]{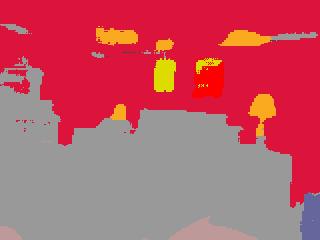} &
    \includegraphics[width=0.14\textwidth]{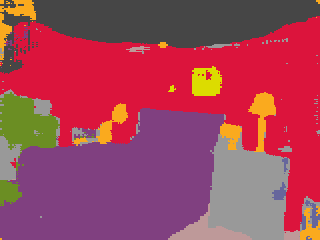} &
    \includegraphics[width=0.14\textwidth]{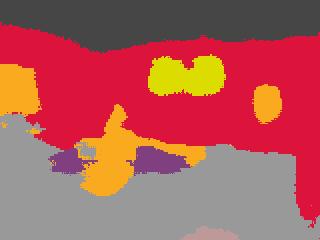} &
    \includegraphics[width=0.14\textwidth]{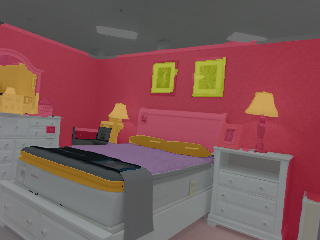} \\
    
    \includegraphics[width=0.14\textwidth]{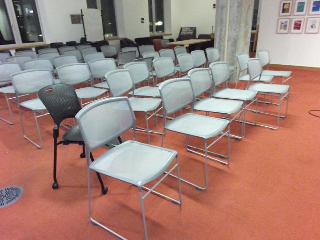} &
    \includegraphics[width=0.14\textwidth]{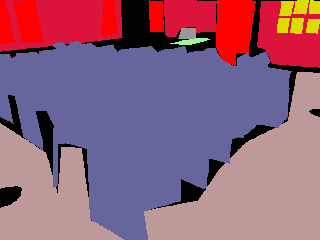} &
    \includegraphics[width=0.14\textwidth]{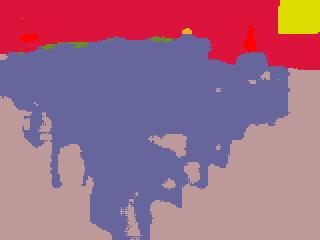} &
    \includegraphics[width=0.14\textwidth]{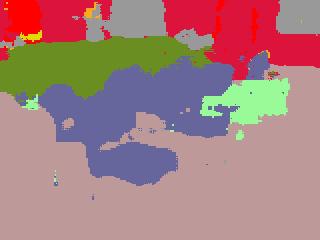} &
    \includegraphics[width=0.14\textwidth]{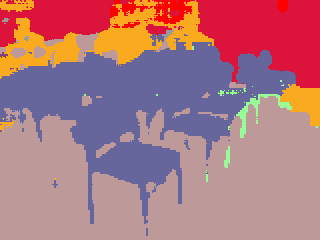} &
    \includegraphics[width=0.14\textwidth]{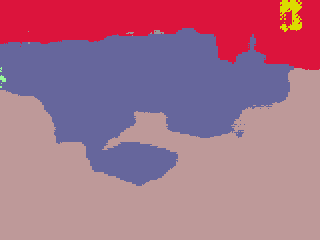} &
    \includegraphics[width=0.14\textwidth]{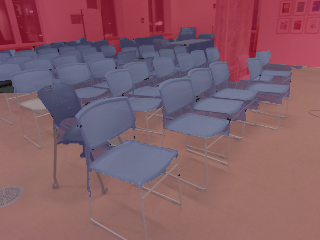} \\
    
    \includegraphics[width=0.14\textwidth]{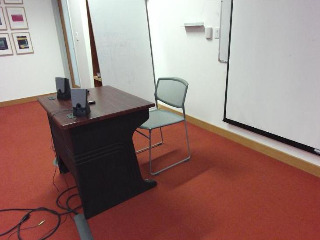} &
    \includegraphics[width=0.14\textwidth]{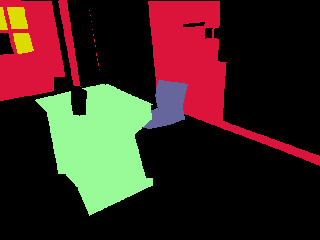} &
    \includegraphics[width=0.14\textwidth]{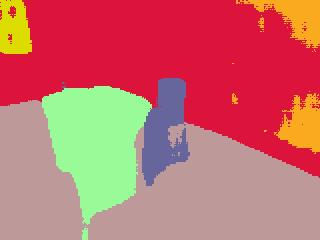} &
    \includegraphics[width=0.14\textwidth]{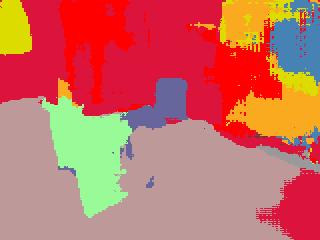} &
    \includegraphics[width=0.14\textwidth]{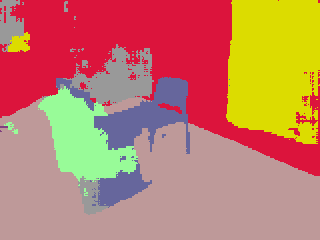} &
    \includegraphics[width=0.14\textwidth]{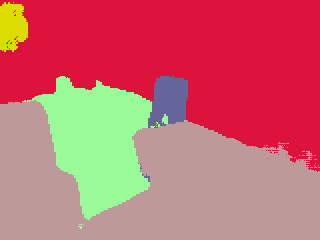} &
    \includegraphics[width=0.14\textwidth]{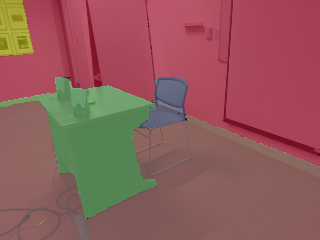} \\
    
    \includegraphics[width=0.14\textwidth]{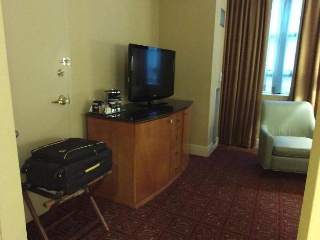} &
    \includegraphics[width=0.14\textwidth]{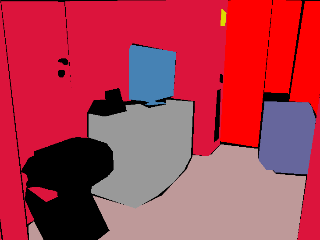} &
    \includegraphics[width=0.14\textwidth]{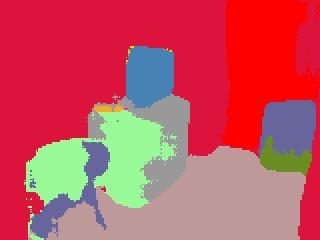} &
    \includegraphics[width=0.14\textwidth]{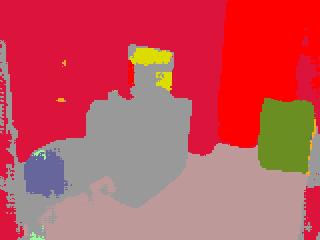} &
    \includegraphics[width=0.14\textwidth]{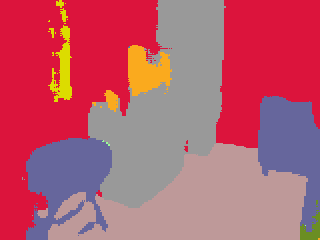} &
    \includegraphics[width=0.14\textwidth]{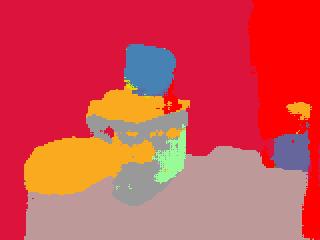} &
    \includegraphics[width=0.14\textwidth]{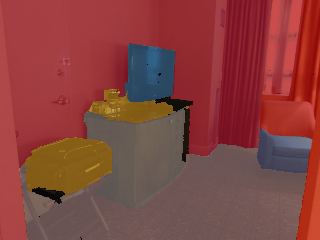} \\
    
    \includegraphics[width=0.14\textwidth]{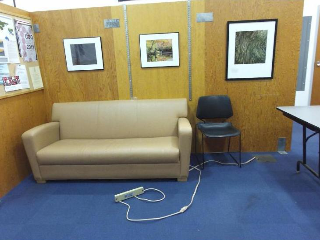} &
    \includegraphics[width=0.14\textwidth]{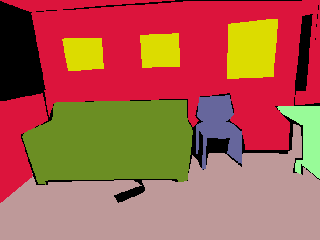} &
    \includegraphics[width=0.14\textwidth]{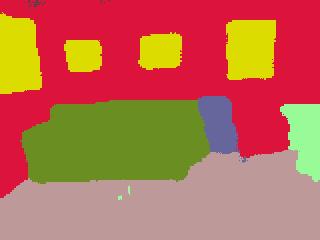} &
    \includegraphics[width=0.14\textwidth]{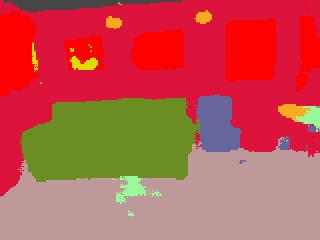} &
    \includegraphics[width=0.14\textwidth]{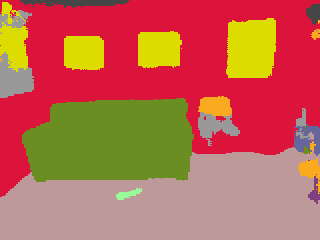} &
    \includegraphics[width=0.14\textwidth]{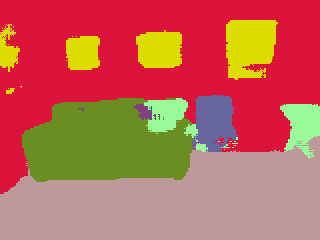} &
    \includegraphics[width=0.14\textwidth]{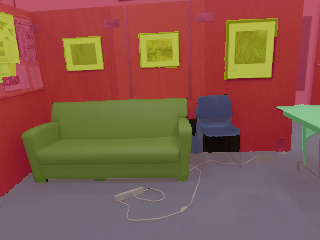}
    
  \end{tabular}
\end{adjustbox}
\end{figure*}

\subsection{Training in Target Domain}
After computing $Y_{\text{Pseudo}}$ as described in the previous stage, the last stage trains the student network $\mathit{SP}_{\text{full}}$ using the real RGB images $X_{\text{Real,RGB}}$ and the estimated labels $Y_{\text{Pseudo}}$. Those estimated labels provide information for only a subset of all pixels, i.e those pixels that we are confident about. In \cite{TonioniPMS17} the authors note that they achieved a better disparity map for whole image with only a portion of high-confidence predictions. Hence, assuming that the majority of estimates in $Y_{\text{Pseudo}}$ are correct, we expect the missing or incorrectly labeled regions to be recovered by the generalization capability of the neural network. Formally, the loss for the student network for a pixel $i$ is given by
\begin{dmath}
L_{\text{Parse},i} = -\sum_{c\in C } w_{\text{Pseudo},c}y_{\text{Pseudo},i,c}\log\left(\frac{e^{p_{i,c}}}{\sum _{c'\in C}e^{p_{i,c'}}}\right)
\end{dmath}
where the pixels in $Y_{\text{Pseudo}}$ with unknown or ignored labels do not contribute to the loss. $p_{i,c}$ denotes the prediction for class $c$ at pixel $i$ and $y_{\text{Pseudo},i,c}$ is a one-hot vector containing the pseudo labels. 

\section{Experiments}

To demonstrate the efficacy of our method, several ablation studies on depth-aware adaptation and cue integration are presented. We evaluate our approach on the SUN RGB-D dataset \cite{SongLX15}. At first, we present experiments to demonstrate the effectiveness of our depth adaptation method. Two synthetic datasets, SceneNet \cite{McCormacHLD16} and Pbrs \cite{ZhangSYSLJF17}, are used during training procedure. Afterwards we proceed to ablation studies of our model to show the effect of each measure on the final result. Finally, we compare to state-of-the-art domain adaptation scene parsing methods and their fully-supervised counterpart. Note that, to be more realistic, no additional real data is used. The only annotations used in our setting are image-level tags from the SUN RGB-D dataset which are much cheaper to acquire than pixel-wise annotations or object bounding boxes.

\subsection{Implementation Details}

All experiments are implemented in the Pytorch 0.3 \cite{paszke2017automatic} framework with CUDA 9.0 and CuDNN backends on a single NVIDIA Titan X. For a fair comparison and consideration of computational efficiency, we evaluate our approaches and the state-of-the-art adaptation method CYCADA \cite{HoffmanTPZISED18} using the ERFNet \cite{RomeraABA18} network architecture. Without loss of generality, our method can be applied to other scene parsing models. Our reproduction of CYCADA is trained with the hyperparameters as published by the authors, including weight sharing.
For the scene parsing model, the input images were resized to $320\times 240$ and the Adam \cite{KingmaB14} variant of stochastic gradient descent is used for minimization of all loss functions. Training is performed with a batch size of 48. Moreover, we train with an initial learning rate of $5\times 10^{-4}$ and reduce it by half once loss value stalls so as to accelerate convergence as done in \cite{RomeraABA18}. We apply standard data augmentation techniques like dropout, random flipping and cropping to prevent our models from overfitting. For the weakly supervised model, we use the encoder of ERFNet pretrained on ImageNet \cite{DengDSLL009} for initialization and replace the original fully-connected layers with a max-pooling, a global average pooling and a new fully-connected softmax layer.

\subsection{Ablation Studies}

\paragraph{Minmax adversarial loss} Table~\textcolor{red}{\ref{tab:normcomp}} demonstrates the effects of minmax normalization on sensor noise learning. The IoU of most categories is improved significantly in the minmax normalization setting over raw depth data. This shows the utility of the normaliser $\eta$ in suppressing depth magnitude based learning in the discriminators $D_N$ and $D_R$.  Note that the loss of category “window” is set to zero and excluded in this comparison due to wrong depth reported by the active sensors employed in creating the SUN RGB-D dataset.

\paragraph{Sensor depth simulation} To show the efficacy of sensor depth simulation, we evaluate the performance of depth-based scene parsing as shown in Table~\textcolor{red}{\ref{tab:mixcomp}}. Our evaluation includes both the raw as well as the inpainted depth maps as provided by the SUN RGB-D. We compare with models trained solely from synthetic data as well as to the simulation method proposed in \cite{BohgRHS14}. Our method clearly outperforms those two methods for both kinds of depth maps, thereby establishing a new baseline for our following adaptation experiments. 

\paragraph{Cues and Integration} Table~\textcolor{red}{\ref{tab:ablation}} disentangles the influence of individual cues and integration mechanisms in each voting stage during training. The results show how both integration stages contribute to the improvement of IoU in different categories to various extent, thereby complementing each other. Class heat maps are particularly helpful for those objects that are smaller or do not possess a distinctive geometric structure. Note that while the overall mIoU performance improves, the mIoU for some categories such as bed and table degrades after the second integration step due to the inter-class occurrence issue: In weak localization where no explicit object positions are available, those categories that mostly appear together in a scene cannot always be properly separated, i.e. their labels could be swapped without invalidating any data.

\paragraph{Student Network} Table~\textcolor{red}{\ref{tab:compst}} compares the result of the student network with the pseudo labels $Y_{\text{Pseudo}}$, i.e. evaluating on $Y_{\text{Pseudo}}$ directly without training the student network. This illustrates how the student network is able to learn a scene parsing model that is more accurate than its training data. The quantities labelled “effective” in the table refers to multiplying the original evaluation matrix with cover ratio, i.e., percentage of valid pixels. Note that effective mIoU is calculated by using class-wise cover ratio instead of global ones.

\subsection{Comparison to the State-of-the-Art}

In Table \textcolor{red}{\ref{tab:compalg}}, we compare our results to full supervision and the state-of-the-art domain adaptation methods. For a fair comparison, all models, including CYCADA are trained using the ERFNet architecture. The not adapting alternative, denoted NADA, is trained on synthetic data directly. CYCADA, the state-of-the-art domain adaption method, was trained starting from the pretrained NADA parameters. Although CYCADA outperforms NADA and performs better than our depth adaptation on categories with indistinctive geometric structures such as “paint”, “tv”, “windows”, there is only a slight improvement, which comes at the high effort of computer generated imagery. It shows that taking appearance from real data into account yields significant advantages even if only image-level labels are available. Fig.~\textcolor{red}{ \ref{fig:finalcomp}} shows examples of our final result. Note that some visualization of our results seems to be incorrect when compared to the ground truth. However, we observed that some ground truths are imprecise and a portion of regions marked "unknown" can be predicted correctly if we align our result with RGB inputs by applying simple UCM based contour-wise voting using predictions at inference phase. Hence, we argue that the performance of our approach may still be underestimated by this evaluation. We provide further examples for this phenomenon in the supplementary material.

\section{Conclusions}
Starting out from synthetically generated scene parsing data, we have demonstrated how transferring information in the depth domain can exploit the smaller domain gap of geometric data for indoor scene parsing. Proceeding to integrate weak localization can recover information that is not present or difficult to detect in synthetic indoor scenery. Altogether this yields a significant performance improvement for learning indoor scene parsing without dense labels, reducing the mIoU drop from 47\% to 20\%. While we utilize depth for our adaptation, this is only necessary at training, not at inference time, thereby maintaining a low computations and sensory footprint. These techniques may readily applied and extended to benefit other computer vision tasks in the future.

{\small
\bibliographystyle{ieee}
\bibliography{ms}
}

\end{document}